\documentclass[10pt,twocolumn,letterpaper]{article}

\usepackage{iccv}
\usepackage{times}
\usepackage{epsfig}
\usepackage{graphicx}
\usepackage{amsmath}
\usepackage{amssymb}

\usepackage{bm}
\usepackage{booktabs}
\usepackage{multirow}
\usepackage[table]{xcolor}

\usepackage{caption}

\usepackage[breaklinks=true,bookmarks=false]{hyperref}

\iccvfinalcopy 

\def\iccvPaperID{3060} 

\ificcvfinal\fi

\def\wrt{{w}\onedot{r}\onedot{t}\onedot}
\def\etal{\emph{et~al}\onedot}

\def\thatis{\emph{i.e}\onedot}

\def\secmk{Sec.~}
\def\figmk{Fig.~}
\def\tablemk{Tab.~}
\def\equationmk{Eqn.~}

\def\boennum{16~}
\def\netname{RUFormer~}

\begin{document}

\title{Nonrigid Object Contact Estimation With Regional Unwrapping Transformer}

\author{
  Wei Xie\hspace{2mm}\hspace{5mm} 
  Zimeng Zhao\hspace{2mm}\hspace{5mm} 
  Shiying Li\hspace{2mm}\hspace{5mm} 
  Binghui Zuo\hspace{2mm}\hspace{5mm} 
  Yangang Wang\footnotemark[1]\\%
\\
Southeast University, China\\
}

\maketitle
\ificcvfinal\thispagestyle{empty}\fi

\begin{abstract}
  Acquiring contact patterns between hands and nonrigid objects is a common concern in the vision and robotics community. However, existing learning-based methods focus more on contact with rigid ones from monocular images. When adopting them for nonrigid contact, a major problem is that the existing contact representation is restricted by the geometry of the object. Consequently, contact neighborhoods are stored in an unordered manner and contact features are difficult to align with image cues. At the core of our approach lies a novel hand-object contact representation called RUPs (Region Unwrapping Profiles), which unwrap the roughly estimated hand-object surfaces as multiple high-resolution 2D regional profiles. The region grouping strategy is consistent with the hand kinematic bone division because they are the primitive initiators for a composite contact pattern. Based on this representation, our Regional Unwrapping Transformer (RUFormer) learns the correlation priors across regions from monocular inputs and predicts corresponding contact and deformed transformations. Our experiments demonstrate that the proposed framework can robustly estimate the deformed degrees and deformed transformations, which makes it suitable for both nonrigid and rigid contact.
  
\end{abstract}

\begin{figure}[!t]
  \centering
  \includegraphics[width=\linewidth]{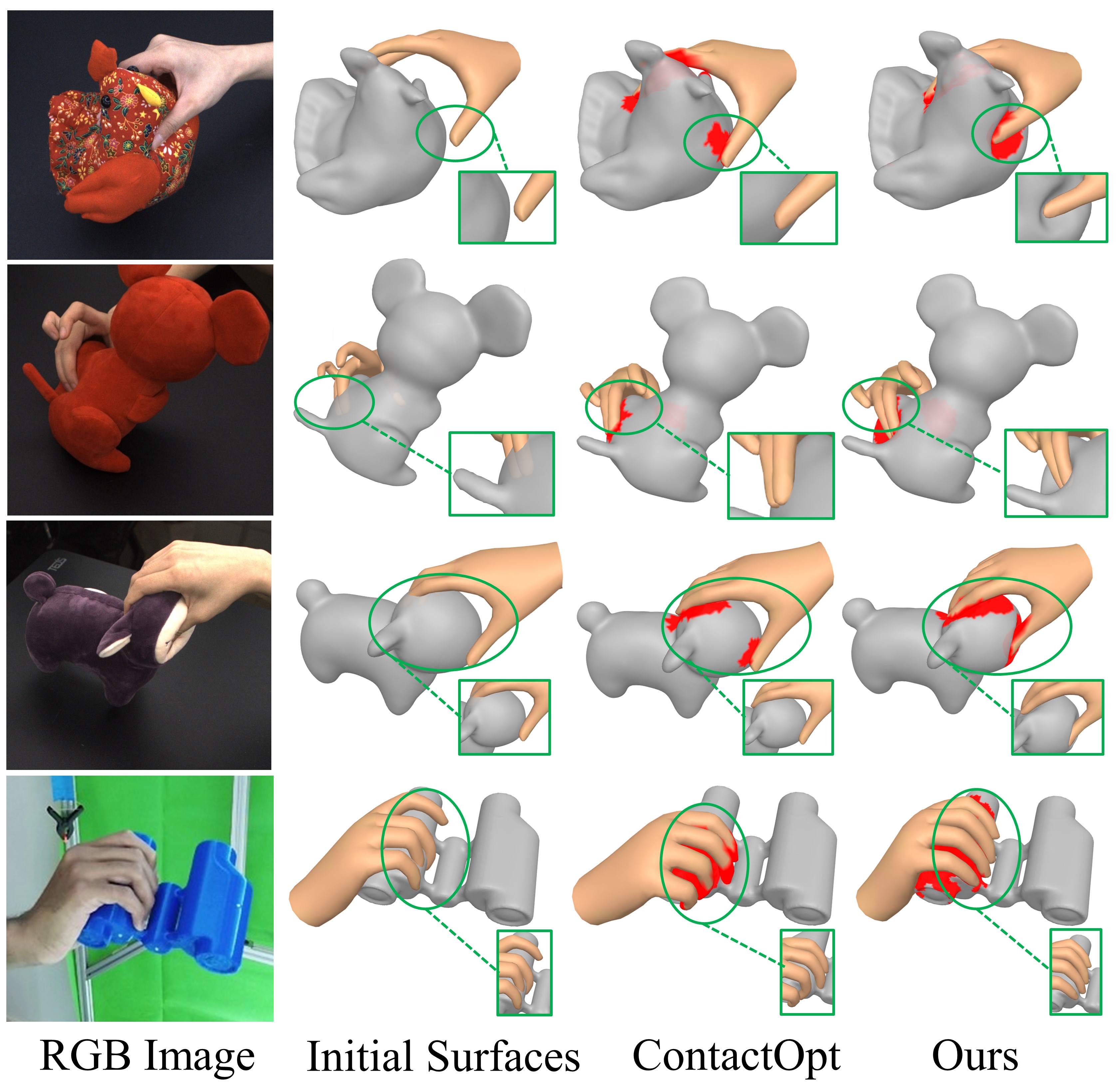}
  \vspace{-4mm}
  \caption{\textbf{Contact patterns estimated from monocular RGB images.}  Since the deformed degrees of the contact areas are considered by our framework, both contact with nonrigid (Row1, Row2, Row3) and rigid objects (Row4) can be plausibly estimated.}
  \vspace{-4mm}
  \label{fig02_teaser}
\end{figure}


\renewcommand{\thefootnote}{\fnsymbol{footnote}}
\footnotetext[1]{Corresponding author. E-mail: yangangwang@seu.edu.cn. All the authors from Southeast University are affiliated with the Key Laboratory of Measurement and Control of Complex Systems of Engineering, Ministry of Education, Nanjing, China. This work was supported in part by the National Natural Science Foundation of China (No. 62076061), the Natural Science Foundation of Jiangsu Province (No. BK20220127).}

\section{Introduction} 

Perceptions of hand-object contact patterns are crucial to advance human-computer interaction and robotic imitation~\cite{zhang2018deep}. The interactive objects in these applications, from mouse/keyboard to bottle/doll, are mostly nonrigid. Although impressive progress has been achieved towards monocular contact estimation between hands and 3D rigid objects~\cite{grady2021contactopt,yang2021cpf,zhao2022stability,tse2022s} or 2.5D cloth~\cite{tsoli2018joint,tsoli2019patch,aranda2020monocular}, it is still difficult to extend them to 3D nonrigid object. One important reason is that existing methods usually project the contact area of different objects onto their own surface (point cloud or mesh), which is represented by either unordered points or unregistered points and edges. As a result, it is challenging to store contact into a feature-aligned space.

\begin{figure}[!t]
    \centering
    \includegraphics[width=\linewidth]{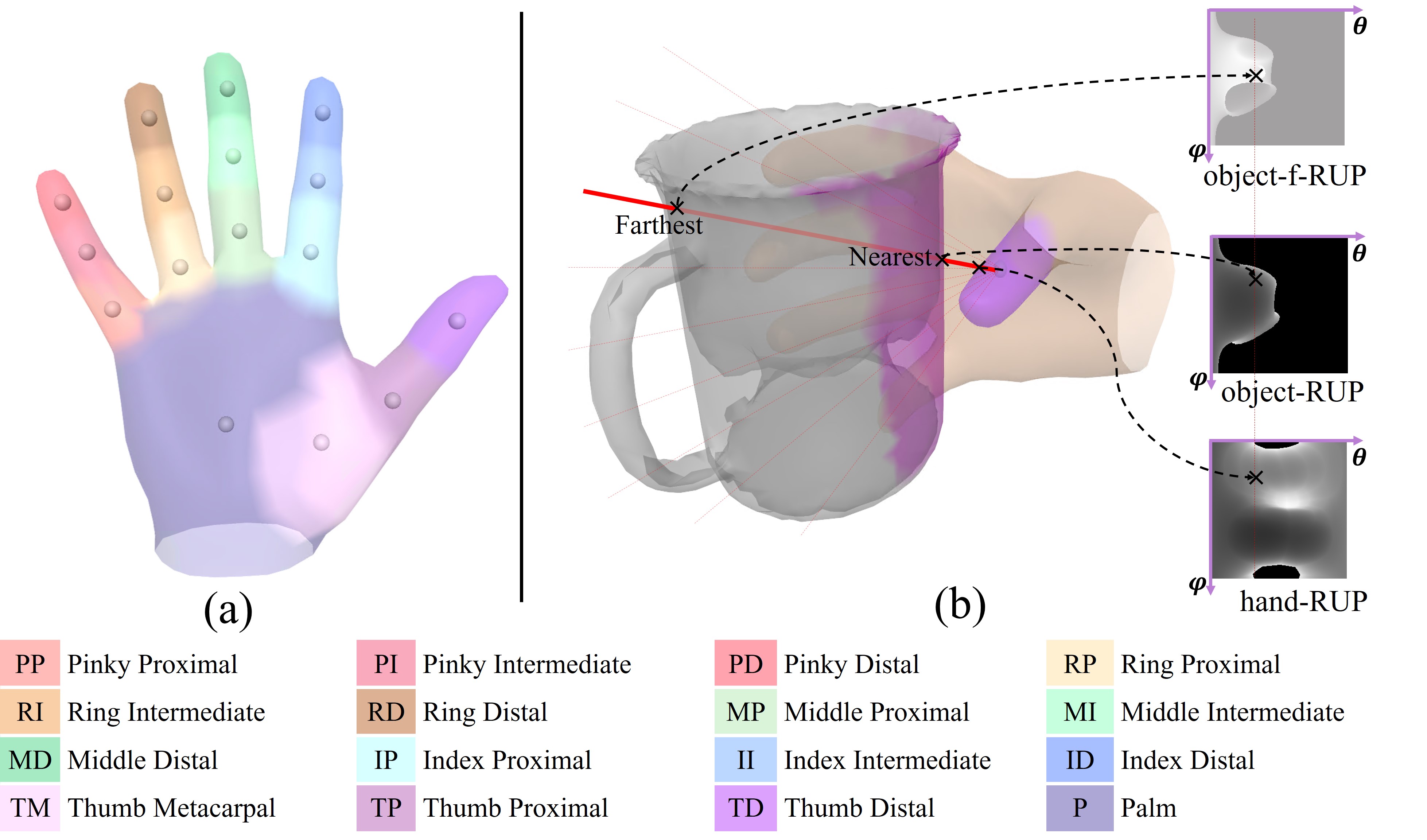}
    \caption{\textbf{Surface grouping strategy.} (a) Hand region division based on 16 kinematic bones of an LBS hand. Each region center is marked as a gray sphere. (b) The correlated hand-object sub-surfaces for each region are aligned according to rays emanating from the center and unwrapped to the 2D regional profiles (\thatis hand-RUP and object-RUP). An extra object-f-RUP is generated only for grid-wise sampling. }      
    \vspace{-4mm}
    \label{fig04_subregions}
\end{figure}

To conquer this obstacle, our key idea is to \textbf{first represent regional 3D surface where hand-object contact may occur as regional 2D unwrapping profiles, then predict the nonrigid contact and deformation within/across regions according to monocular image cues through a Vision Transformer}. Considering that the mutual contact is caused by individual hand regions~\cite{hasson2020leveraging,yang2021cpf,zhao2022stability}, our surface grouping is based on the \boennum hand kinematic bones~\cite{romero2017embodied,mihajlovic2021leap,yang2021cpf,zhao2022stability} illustrated in \figmk\ref{fig04_subregions}(a). Each piece of object surface shown in \figmk\ref{fig04_subregions}(b) is divided into a certain group when it can be directly intersected by a ray emanating from the region center associated with this group. Each subsurface is further mapped to the image plane according to the spherical unwrapping algorithm~\cite{zhao2021supple}. Consequently, the whole object surface is converted to \boennum object \emph{regional unwrapping profile}s (object-RUPs). Similarly, the hand surface is converted to \boennum hand-RUPs, each of which records pixel-aligned ray intersections with the object-RUP in the same group. In contrast to object point clouds~\cite{karunratanakul2020grasping,grady2021contactopt,tse2022s,chen2022tracking}, this novel representation preserves both the hand-object surface correlation and the contact point orderliness.  

Numerous works~\cite{grady2021contactopt,tse2022s} only predicted plausible contact patterns according to data prior and ignore contact clues in the image. This may be applicable to rigid interaction. However, when the deformed degree is considered, multiple nonrigid contact patterns can be yielded from the same hand-object spatial relationship. Therefore, our framework crops the image patches of the corresponding \boennum hand bones as extra visual cues to estimate nonrigid contact. Altogether, our \netname is tamed to take those \boennum groups of hand-RUPs, object-RUPs, and visual cues as the inputs. It gradually estimates the contact and deformed features across RUPs, and finally predicts the deformed transformations of the object. To our best knowledge, this is the first framework that is applicable to reconstruct both rigid and nonrigid hand-object interaction from monocular images.     

In summary, our main contributions are:

\noindent$\bullet$ A learning-based framework with the ambition to estimate the contact between hand and nonrigid objects from monocular images; 

\noindent$\bullet$ A hand-object interaction representation to record hand-object surfaces into multiple pixel-aligned and fine-grained 2D profiles; 

\noindent$\bullet$ A unwrapping-informed transformer to predict contact and deformation on the object according to both visual cues and data prior.

\section{Related Work} 

\noindent\textbf{Hand-object interaction reconstruction.} 
Thanks to the creation of several hand-object interaction datasets~\cite{hampali2020honnotate, chao2021dexycb, zhao2022stability, brahmbhatt2020contactpose,hampali2021handsformer,kwon2021h2o,taheri2020grab} in recent years, monocular 3D hand-object interaction reconstruction has received extensive attention from researchers. Hasson~\etal~\cite{hasson2019learning} proposed a two-branch network to reconstruct the hand and an unknown manipulated object. Subsequent works\cite{chen2022alignsdf,ye2022s} estimated hand-object pose and inferred implicit 3D shape of the object. 
Other works~\cite{tekin2019h+,doosti2020hope,hasson2020leveraging,hasson2021towards,cao2021reconstructing,yang2021cpf,zhao2022stability} assumed that the object template is known and reduce the object reconstruction to 6D pose estimation. They jointly regressed hand and object poses by reasoning about their interactions. 
However, all the existing work focuses on interactions between hands and rigid objects. Our framework attempts for the first time to reconstruct the interaction between hands and non-rigid objects from monocular images, while also being compatible with rigid ones. 

\begin{figure*}[!t]
    \centering
    \includegraphics[width=\linewidth]{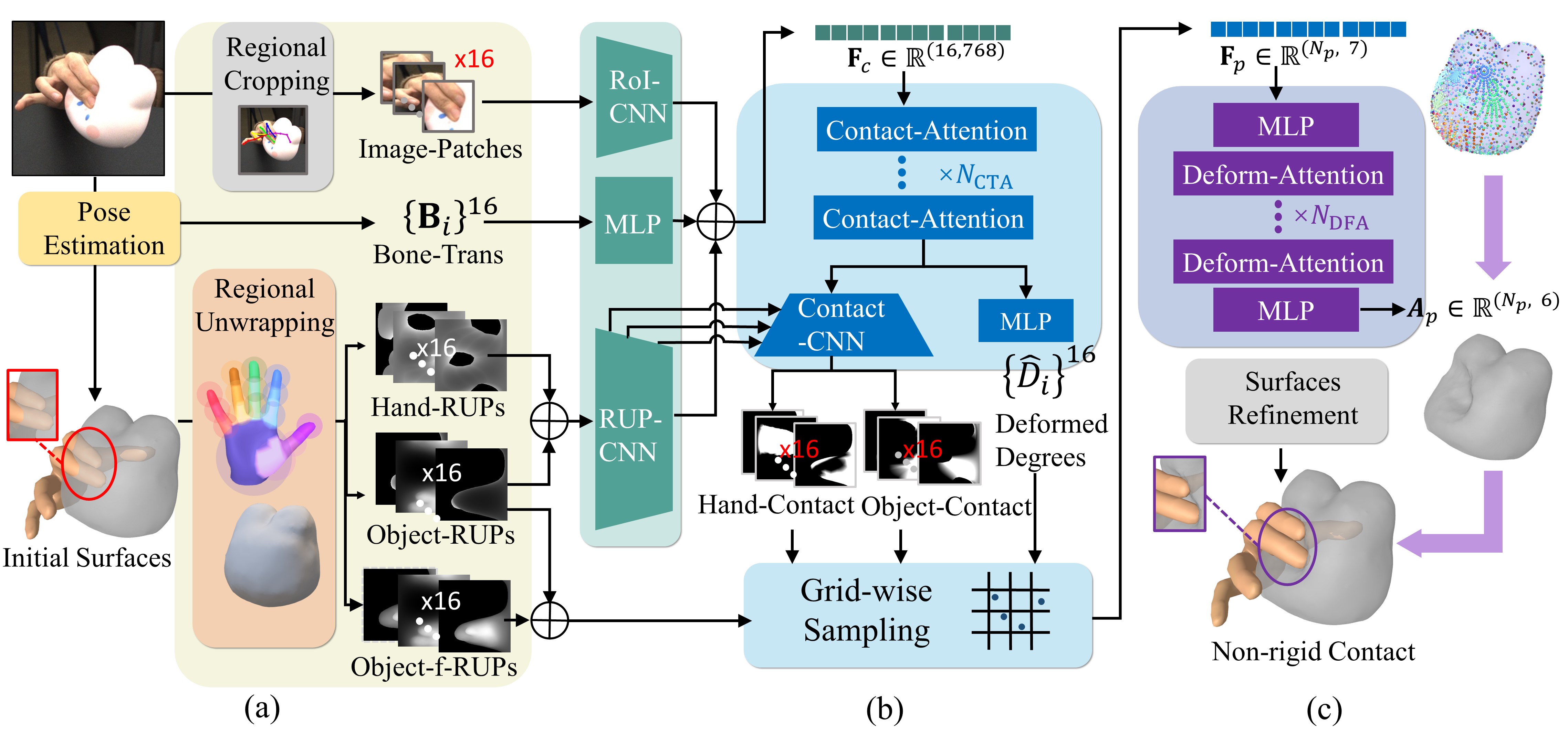}
    \vspace{-4mm}
    \caption{\textbf{Overview of \netname.} (a) The preparation process of \netname input data, all of which are aligned to the hand 16 regions. (b) \netname Encoder estimates hand-object regional contact areas from image patches, hand bone transformation and RUPs. (c) \netname Decoder estimates fine-grained deformation from grid-wise sampling features.
    }
    \vspace{-2mm}
    \label{fig02_pipeline}
\end{figure*}

\noindent\textbf{Hand-object contact pattern estimation.}
Inferring contact patterns is vital for 3D hand-object reconstruction. \cite{hasson2019learning, cao2021reconstructing} introduced contact losses which encourage contact surfaces and penalize penetrations between hand and object. However, these methods cannot enforce hand-object alignment at test time. 
Recently, Some works~\cite{grady2021contactopt,tse2022s,yu2022uv} used explicit contact inference and enforcement to achieve higher quality grasps.
Grady~\etal~\cite{grady2021contactopt} estimated the contact pattern between hand and object based on PointNet~\cite{qi2017pointnet}. Tse~\etal~\cite{tse2022s} proposed a graph-based network to infer contact patterns. \cite{grady2021contactopt,tse2022s} estimated hand-object contact patterns from sparse point clouds, which are unordered and challenging to store contact into a feature-aligned space. Yu~\etal~\cite{yu2022uv} proposed a dense representation in the form of a UV coordinate map, which only inferred the contact areas of the hand surface. All the existing works focus on contact with rigid objects and are not applicable to contact with non-rigid objects. 


\noindent\textbf{Vision transformer.} 
Transformer and self-attention networks have revolutionized natural language processing~\cite{vaswani2017attention,devlin2018bert, wu2019pay} and are making a deep impression on visual-related tasks, such as object detection~\cite{carion2020end,zhu2020deformable}, image classification~\cite{dosovitskiy2020image}, 3D pose estimation~\cite{huang2020hand,he2020epipolar,lin2021end, huang2022occluded, hampali2022keypoint} and point cloud processing~\cite{tang2022neural}. We refer the reader to~\cite{han2020survey} for a details survey of Vision Transformer. In our task, We use attention modules to exploit the visual and hand-object spatial correlations.


\section{Method} 
An overview of our pipeline for hand and nonrigid objects contact patterns estimation is shown in~\figmk\ref{fig02_pipeline}. It takes the image and the corresponding object mesh template as input. Through our \netname, it predicts the contact areas of the hand-object surface pair, as well as the deformation of the object. Initially, the hand-object surfaces are estimated and unwrapped into multiple high-resolution 2D regional profiles (\secmk\ref{sec32_region}). Then, \netname estimates contact according to region-aligned features (\secmk\ref{sec33_contact}), and predicts deformed transformations of sampling points (\secmk\ref{sec34_deform}). 
Hand-object surfaces refinement and deployment details are described in \secmk\ref{sec35_implementation}.

\subsection{Preliminary}
\label{sec31_preliminary}
\noindent\textbf{Surface representation.} 
We represent the hand surface based on MANO~\cite{romero2017embodied}. It can be regarded as a differentiable function $M_{h}(\bm{\beta},\bm{\theta},\bm{\tau})$ parameterized by shape $\bm{\beta}\in \mathbb{R}^{10}$, pose $\bm{\theta}\in \mathbb{R}^{16\times3}$ and global translation $\bm{\tau}\in \mathbb{R}^{3}$ \wrt the camera coordinate system. For a left-hand case, the RGB images and hand-object surfaces are mirrored together in advance. ~~We represent object 6D pose as its mesh template \wrt the right MANO hand coordinate system. It is noted that the plausible hand-object relationship is always the object in front of the hand, \thatis $y<0 $ in the vertex coordinates of the object. The deformation of a sampling point $\bm{p}$ on the nonrigid object is represented as an affine transformation $ \mathbf{A}(\bm{p}) \in SE(3)$ \wrt to its template position. 

\noindent\textbf{Grouping strategy.} 
Our hand-object surfaces grouping is shown in~\figmk\ref{fig04_subregions}. We first divide the hand region based on the 16 kinematic bones of the posed MANO. Each piece of the object surface is divided into a certain group when it can be directly intersected by a ray emanating from the hand region center associated with this group. We further unwrap each subsurface to the image plane to obtain \boennum hand-RUPs, \boennum object-RUPs and \boennum object-f-RUPs. The records are pixel-aligned in the same group. It should be noted that object-f-RUPs are only used for grid-wise sampling.


\subsection{\netname Input: Region Alignment}
\label{sec32_region}
\noindent\textbf{Surface initialization.} 
The hand-object pose estimation network is refer to~\cite{zhao2022stability}. We take RGB image and object template as inputs and predict the MANO parameters and object 6DoF pose. We integrate MANO as a differentiable network layer and use it to output the 3D hand surface. 


\begin{figure}[!t]
    \centering
    \includegraphics[width=\linewidth]{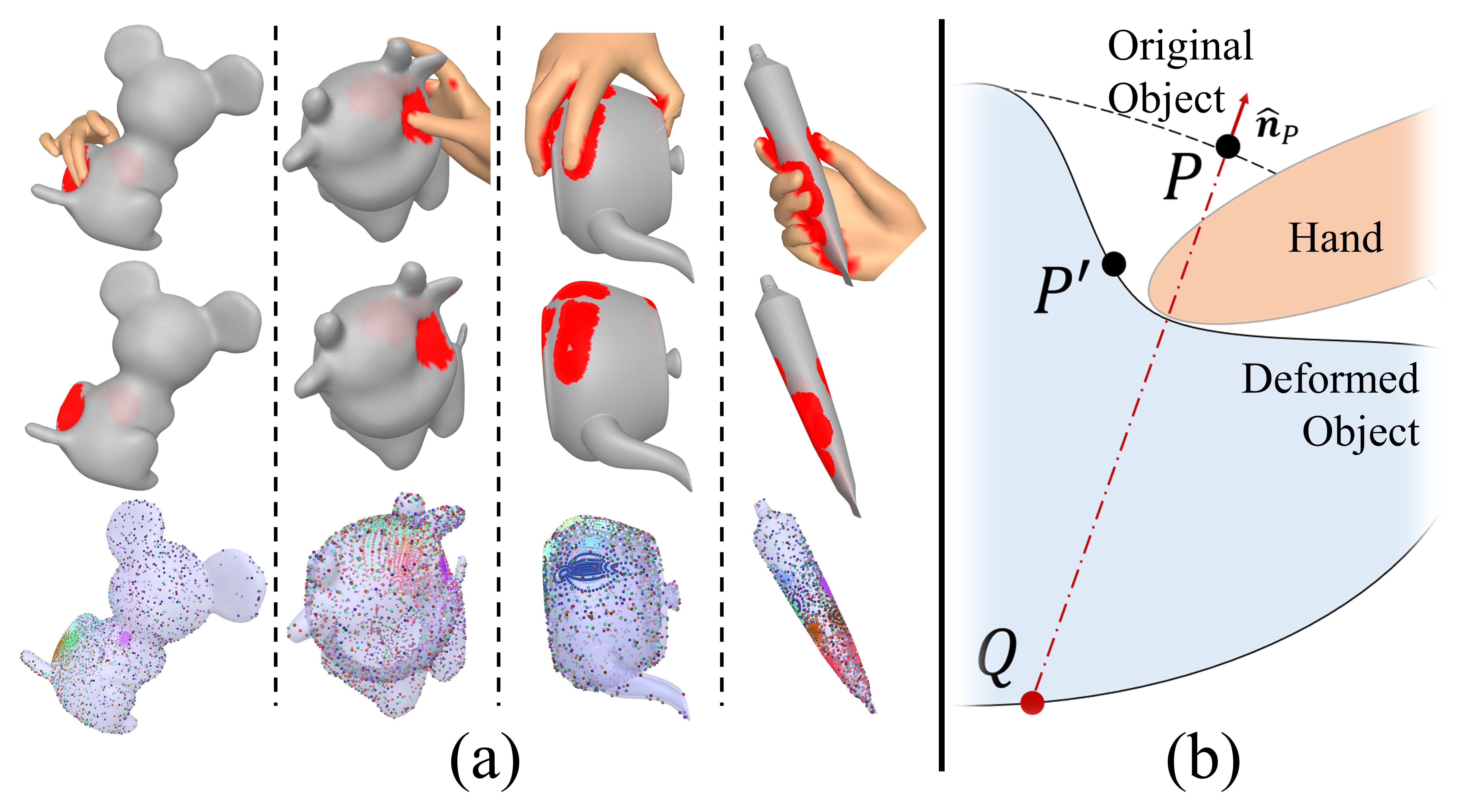}
    \vspace{-3mm}
    \caption{\textbf{(a) Grid-wise sampling} results after back-projecting on their object surfaces. Each column corresponds to an instance. \textbf{(b) The illustration of object deformation process.} The \emph{deformation vector} of a surface point $P$ is defined as its position difference before and after deformation ($PP'$). The \emph{maximum deformation} of $P$ is defined as the distance $d_{PQ}$ between the point in its original state and the closest intersection point of the object projected from that point in the opposite normal direction. The \emph{deformation degree} of $P$ is $\bm{v}_P \triangleq d_{PP'} / d_{PQ}$.
    }
    \vspace{-3mm}
    \label{fig03_gridsampling}
\end{figure}

\noindent\textbf{Surface unwrapping.} We unwrap the estimated hand-object surfaces into multiple fine-grained RUPs. RUPs define the projection to unwrap the hand-object surfaces into the image plane with the center of 16 hand sub-surfaces as the origin, respectively. 
We refer to~\cite{zhao2021supple} to map a surface point $\bm{p}(x, y, z)$ in the Cartesian coordinate system to the spherical coordinate $\bm{s}(\rho, \theta, \varphi)$. 
Specifically, the closest intersections between the hand surface and the rays emitted from $i$-th bone center are recorded in the $i$-th hand-RUPs, the closest intersections between the object surface and those same rays are recorded in the $i$-th object-RUPs, and the farthest intersections between the object surface and those same rays are recorded in the $i$-th object-f-RUPs. All rays emitted from a center can be parameterized as $\overrightarrow{O_i R}(\vartheta, \varphi)$, where $\vartheta \in [0, \pi], \varphi\in [0, 2\pi], \rho >0 $ are the spherical coordinates. Therefore, each RUP channel can be formulated as: 
\begin{equation}
    \mathbf{R}\left(\frac{\vartheta}{\pi} W_{R}, \frac{\varphi}{2 \pi} H_{R}\right) \triangleq \underset{\rho}{\arg \min }\{\bm{s}(\rho) \mid \bm{s} \in(\overrightarrow{O_i R} \cap \partial \bm{S})\}
    \label{eqn_rup}
\end{equation}
where $\mathbf{R}$ denotes as RUP and $\partial \bm{S}$ represents hand or object surface. The value $\rho$ is set to zero if no interaction occurs. As a result, hand-RUPs $\{\mathbf{R}_i^{H} \}_{i=1}^{16}$, object-RUPs $\{\mathbf{R}_i^{O} \}_{i=1}^{16}$ and object-f-RUPs $\{\mathbf{R}_i^{OF} \}_{i=1}^{16}$ are all 16-channel image tensors. Furthermore, pixels with the same indices correspond to the intersections between the same ray and the hand bone surface / near object surface / far object surface. 

\noindent\textbf{Regional features.} 
Our \netname utilizes regional aligned features to predict contact areas, deformed degree of contact areas and deformed transformations of the object.
With the above estimation, the following features are aligned according to the region: (i) The image patches  $\left\{\mathbf{I}_i\right\}_{i=1}^{16}$ belonging to each bone are cropped using the guidance of MANO joint image coordinates, where $\mathbf{I}_i \in \mathbb{R}^{(3,n_p,n_p)}$. (ii) The MANO pose is further converted as bone transformations $\left\{\mathbf{B}_i\right\}_{i=1}^{16}$ to measure the relative relationship across RUP groups, where $\mathbf{B}_i \in SE(3)$. (iii) $\{\mathbf{R}_i^{H} \}_{i=1}^{16}$, $\{\mathbf{R}_i^{O} \}_{i=1}^{16}$ and  $\{\mathbf{R}_i^{OF} \}_{i=1}^{16}$ computed from the  estimated hand-object 3D surface. 

\subsection{\netname Encoder: Contact Estimation}
\label{sec33_contact}

\noindent\textbf{Contact attentions.} 
Our contact area estimation process is shown in \figmk\ref{fig02_pipeline}(b). Image patches, bone transformations, hand-RUPs and object-RUPs are embedded to the latent space through their respective feature extractors. 
We extract regional image features from 16 groups of image patches by the first four blocks of ResNet18~\cite{he2016deep}. The regional unwrapping features are extracted from 16 groups of object-RUPs and hand-RUPs through the first four blocks of ResNet18. The bone transformations are sent to the MLP to encode features of the relative relationship. These features are later concatenated together as a regional feature embedding $\mathbf{F}_c \in \mathbb{R}^{16 \times 768}$. 
After that, $N_{CTA}$ cascaded ViT-based~\cite{dosovitskiy2020image} attention modules are used to exploit the visual and hand-object spatial correlations within/across these 16 groups. It computes contact embeddings $\mathbf{F}_{c+}$ with the same size as $\mathbf{F}_c$.

\noindent\textbf{Contact representation.} 
We represent hand contact as 2D maps $\{\mathbf{C}_i^{H}\}_{i=1}^{16}$ on hand-RUPs and object contact as 2D maps $\{\mathbf{C}_i^{O}\}_{i=1}^{16}$ on object-RUPs. Each pixel on the contact map indicates the contact probability of the point recorded on RUP with the same indices. To estimate these image-like tensors, an extra CNN decoder with a symmetrical structure with RUP encoder is further adopted. The contact embeddings $\mathbf{F}_{c+}$ are up-sample back to the RUP space again. 

\noindent\textbf{Deformed degree.} 
Besides hand-object contact maps, we further estimate regional deformed degree $\{D_i\}_{i=1}^{16}, D_i \in [0,1]$ from $\mathbf{F}_{c+}$ by an MLP. 
Each value in $\{D_i\}_{i=1}^{16}$ represents the deformed degree of the contact area in the 16 object-RUPs. 

\noindent\textbf{Loss terms.} During the training, we supervised the hand-object contact maps and the deformed degree of contact areas. The \netname encoder loss $L_{C}$ can be expressed as follows:
\begin{equation}
    \begin{aligned}
        {L_{C}} =  {L}_{M} + \lambda_{1}  {L}_{D}
    \end{aligned}
    \label{eqn_ConNet}
\end{equation}
where $L_{D}$ is the standard binary cross-entropy loss for deformed degrees, and $L_{D} = 1000$. 
The ground truth of the deformed degree for each RUP is the average deformed degree of all contact points within the region. The process of obtaining the deformation degree of the contact points is shown in \figmk\ref{fig03_gridsampling}(b).
$L_{M}$ is the MSE loss between the prediction and the ground truth of hand-object contact maps, which can be defined as:
\begin{equation}
    \begin{aligned}
        {L}_{M} = \sum_{i=1}^{16} ( \|{\mathbf{C}}_i^{H} - \hat{\mathbf{C}}_i^{H}\|_2^{2} +   \|\mathbf{C}_i^{O} - \hat{\mathbf{C}}_i^{O}\|_2^{2}  )
    \end{aligned}
    \label{eqn_ConNet}
\end{equation}
where $\mathbf{C}_i^{H}$ and $\mathbf{C}_i^{O}$ are ground truth. $\hat{\mathbf{C}}_i^{H}$ and $\hat{\mathbf{C}}_i^{O}$ are predicted contact maps.

\subsection{\netname Decoder: Deformation Estimation}
\label{sec34_deform}
\noindent\textbf{Coarse deformation acquiring.} 
The focus of the previous sections is on the contact area. However, fine-grained deformation should be described from point perspective.
For each pixel on the object RUPs, it corresponds to a point on the object surface. With the help of deformed degrees of contact areas, the coarse deformation of contact points can be obtained. As illustrated in \figmk\ref{fig03_gridsampling}(b), the ray is emitted from the point to the object surface for intersection detection, and its maximum deformation is defined as the distance between the point from the closest intersection point of the object. The ray direction is set to the negative normal direction of the point. The coarse deformation of the contact point is the maximum deformation multiplied by the predicted deformed degree. However, the deformation obtained in this way does not consider the deformation priori of local geometries and lacks the understanding of the global deformation behavior.

\noindent\textbf{Points sampled from RUPs. }Therefore, we sample points from the object and utilize the \netname decoder to aggregate deformation features from these sampling points, ultimately predicting the deformation transformations of the object. Existing practices utilize the farthest point sampling to acquire point candidates, or iteratively optimize them through geodetic distance. By contrast, because the surface points of the object have been divided into 32 groups ($\{\mathbf{R}_i^{O} \}_{i=1}^{16}$ and  $\{\mathbf{R}_i^{OF} \}_{i=1}^{16}$) based on their distance from each hand bone, we select sampling points from RUPs in an orderly manner. Specifically, we divide RUP into $n_g \times n_g$ grids and sample one point within a grid with maximum value. The coordinates of sampled points are converted back to Cartesian coordinates. For a grid with all-zero pixels, we use the one mask embedding $\bm{p}_{[M]} \in \mathbb{R}^3$ ~\cite{he2022masked} as a replacement: 
\begin{equation}
    \bm{p} = \begin{cases}
        \Pi^{-1}(\rho, \theta, \varphi) & \rho \neq 0 \\
    \bm{p}_{[M]} , & \rho = 0 \\
        \end{cases}
\end{equation}
where $\bm{p}$ and $\rho$ are the 3D point and pixel value corresponding to pixel $(\theta,\varphi)$ in a RUP. We obtain ordered point candidates. ~~For a contact point, its deformation feature is set as its coarse deformation. For a non-contact point, its deformation feature is set as the learnable embedding $\bm{d}_{[M]} \in \mathbb{R}^3$.
As shown in ~\figmk\ref{fig03_gridsampling}(a), the points sampled according to RUP grids emphasize more on contact area compared with other general sampling strategies. 
With the above conversion, $\frac{H_R}{n_g} \times \frac{W_R}{n_g} \times 32$ points are sampled.

\noindent\textbf{Deformation attentions.} 
We inherit the idea of the deformation graph~\cite{sumner2007embedded} that represents the deformation of arbitrary points on the surface as a combined deformation of nearby nodes:
\begin{equation}
    \begin{aligned}
        \tilde{\bm{p}} =\sum_{m=1}^k \omega_{m}\left[\mathbf{A}_{m}\left({\bm{p}} -\bm{g}_{m}\right)+\bm{g}_{m}\right]
    \end{aligned}
    \label{eqn_object_deform_mesh}
\end{equation}   
where $\bm{p}$ is original position of the point and $\tilde{\bm{p}}$ is its deformed position. $\omega_{m}$ is the weight of node $\bm{g}_{m}$ to $\bm{p}$. The weight calculation is referred to~\cite{sumner2007embedded}. Therefore, the \netname decoder is designed to select $N_q$ nodes from $N_p$ points, and predict their affine transformations $\{\bm{A}_{k}\}_{k=1}^{N_p} $ according to input features $\mathbf{F}_p$. In practice, we use the farthest point sampling to select $N_q$ nodes from $N_p$ points, where $N_p = \frac{H_R}{n_g} \times \frac{W_R}{n_g} \times 32$. The input features $\mathbf{F}_p = \{\bm{p}_{j} \oplus \bm{d}_{j} \oplus c_j \}_{j=1}^{N_p} \in \mathbb{R}^{N_p\times 7}$, where $c_j$ indicates whether the point is selected as the node. If the point is selected as the node, it is 1, otherwise, it is 0. Our deformation transformations of the object estimation process are shown in~\figmk\ref{fig02_pipeline}(c). We first utilize an MLP to encode $\mathbf{F}_{p}$ to the latent space and extract deformation embeddings $\mathbf{F}_{d} \in \mathbb{R}^{N_p\times 256}$. After that, $N_{\operatorname{DFA}}$ cascaded attention modules are used to enhance the understanding of global deformation behavior and aggregate the deformation features. It computes embeddings $\mathbf{F}_{d+}$ with the same size as $\mathbf{F}_d$. Finally, the deformation transformations are obtained through an MLP. We train RUFormer decoder in a semi-supervised manner (transformations of points not selected as nodes are not supervised). To reduce dimensionality, each rotation is represented as an axis-angle.

\subsection{Implementation Details}
\label{sec35_implementation}
\noindent\textbf{Surface refinement.} 
Based on hand-object contact maps and deformation transformations, we refine the hand and object surfaces. We first perform object surface deformation. The vertices in the object are deformed by~\equationmk\ref{eqn_object_deform_mesh}. Afterward, hand and object pose are refined based on hand contact maps $\{\hat{\mathbf{C}}_i^{H}\}_{i=1}^{16}$ and object contact maps $\{\hat{\mathbf{C}}_i^{O} \}_{i=1}^{16}$. We convert back to points in the surface by querying pixels in hand-RUPs and object-RUPs, then obtain the contact information of the hand and object surface vertices through interpolation, respectively. Then we follow the method in~\cite{grady2021contactopt} and optimize the hand-object poses to achieve the target contact.



\noindent\textbf{Parameter settings.} 
The hand-object RUPs size is set to  $ H_{R} = W_{R} = 64 $ and the image patch size is set to $ n_p = 64 $. The grid size for point sampling is set to $n_g = 4$. The depth of contact attentions modules and deformation attentions modules are set to $N_{\operatorname{CTA}} = 6, N_{\operatorname{DFA}} = 5$, respectively. 
We use Pytorch to implement our networks and train them on a computer configured with NVIDIA GeForce RTX 3090. \netname encoder and \netname decoder are trained separately. To train \netname encoder, we use SGD optimizer with a learning rate 1e-4. \netname decoder is trained with Adam optimizer with a learning rate 1e-4. The total training epochs for \netname encoder and \netname decoder are both 100. 



\section{Experiments}

\begin{figure}[!t]
    \centering
    \includegraphics[width=\linewidth]{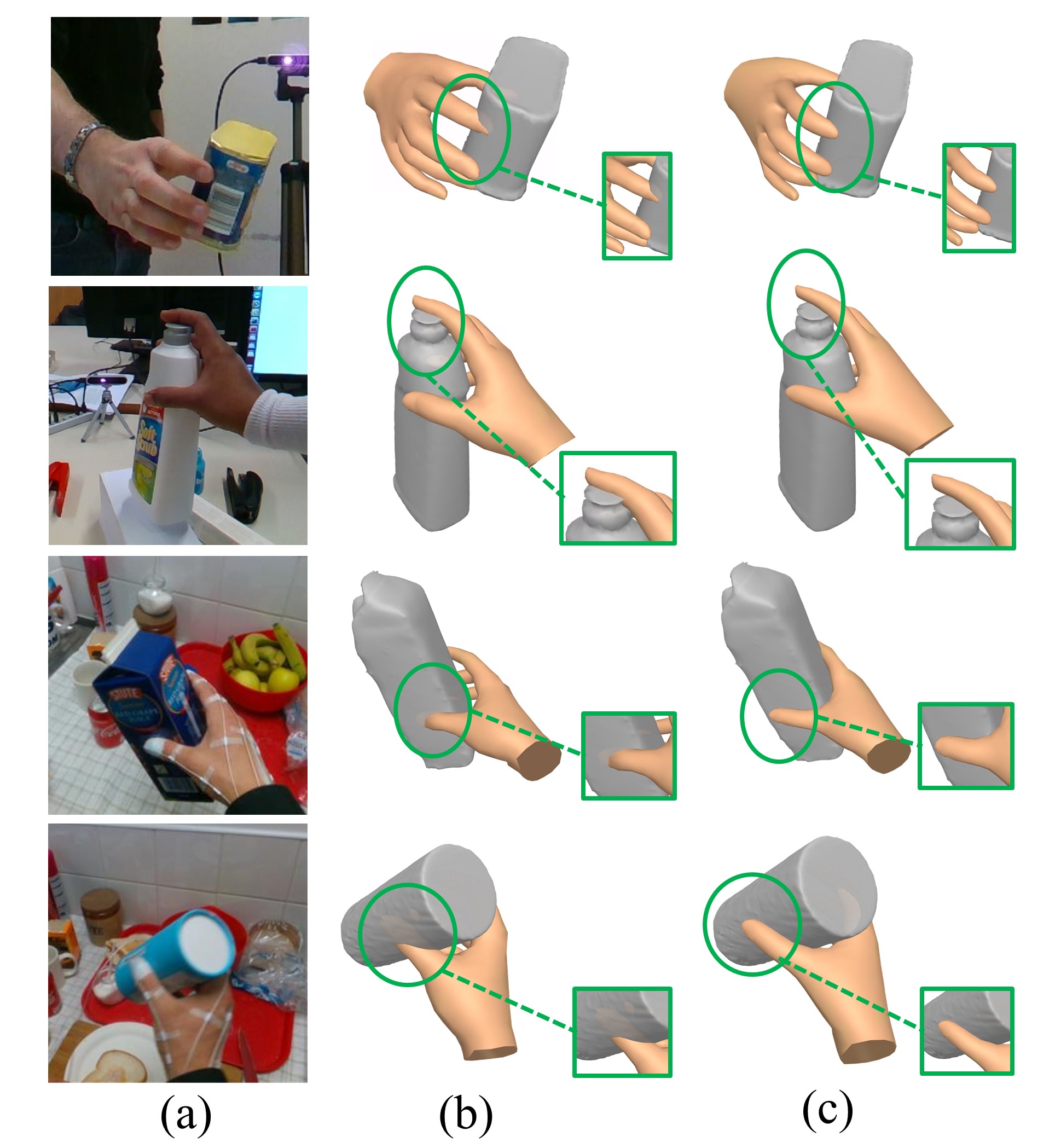}
    \caption{\textbf{Comparisons on monocular reconstruction. } (a) RGB images.  (b) Reconstruction results from~\cite{hasson2020leveraging}. (c) Ours.}
    \vspace{-2mm}
    \label{fig04_mono_rec}
\end{figure}

\begin{figure*}[!t]
    \centering
    \includegraphics[width=\linewidth]{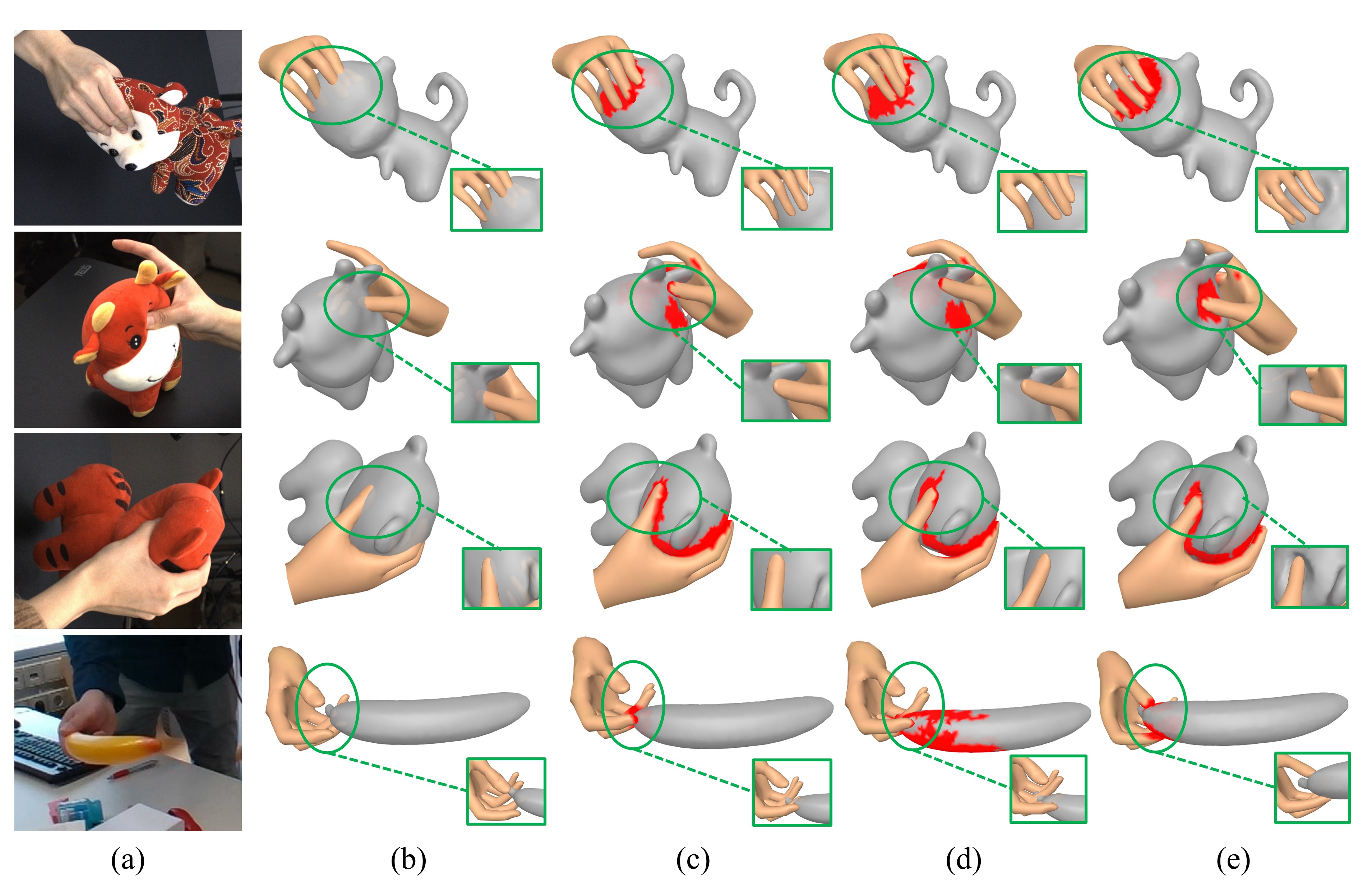}
    \caption{\textbf{Comparisons on contact patterns optimization.} (a) RGB images. (b) Initial hand-object surfaces. (c) Results from ContactOpt~\cite{grady2021contactopt}. (d) Results from $\mathrm{S}^2$Contact~\cite{tse2022s}. (e) Ours. Row1, Row2 and Row 3 are 3D nonrigid interactions. Row4 is 3D rigid interaction.}  
    \vspace{-2mm}
    \label{fig07_contact_driven}
\end{figure*}

\subsection{Datasets}
Our experiments are performed on the HMDO~\cite{xie2023hmdo}, HO3D~\cite{hampali2020honnotate}, FPHB~\cite{garcia2018first} and ContactPose~\cite{brahmbhatt2020contactpose} datasets. The HMDO dataset records the interaction between hands and nonrigid objects. We split it with 4:1 for training and testing. HO3D and FPHB is the dataset of hands in manipulation with rigid objects. We follow the official dataset split for HO3D and adopt the action split following the protocol given by~\cite{hasson2020leveraging} for FPHB. ContactPose is the dataset of hand-object contact paired with hand-object pose. HMDO, HO-3D, and FPHB datasets are used to test our entire pipeline. ContactPose dataset is used to evaluate our hand and object contact estimation. To reduce the ambiguity in the selection of interacted objects, we filter these datasets with the 3D distance between the hand-object not exceeding $2 \mathrm{mm}$ as the threshold. 

\subsection{Metrics}
\noindent\textbf{Hand-object error.} We use the mean per-point position error (\emph{MPJPE}) of 21 hand joints to evaluate the 3D reconstruction error. The mean per-vertex position error (\emph{MPVPE}) is adopted to evaluate the object error.

\noindent\textbf{Contact quality.} We adopt \emph{max penetration} (denoted as Max Pene. in tables) and \emph{intersection volume} (denoted as Inter. in tables) proposed in~\cite{hasson2019learning} to evaluate the hand-object geometric relationship.

\begin{table}[t]
    \begin{center}
        \resizebox{1\linewidth}{!}{
    \begin{tabular}{c|cccc}
    \noalign{\hrule height 1.5pt}
    \rowcolor{white}Methods 
    & Initial State
    &~\cite{grady2021contactopt} &~\cite{tse2022s} &Ours
    \\  
    \midrule
    $\text{MPJPE}_{H}(\mathrm{mm})\downarrow$
    &14.67  &14.92 &14.81 &\textbf{14.78}    
        \\  
    Max Pene.$(\mathrm{mm})\downarrow$ 
    &11.82   &9.46 &9.75 &\textbf{9.24}    
         \\  
    Inter.$(\mathrm{cm}^3)\downarrow$
    &10.69   &7.51 &7.58 &\textbf{7.39}     
        \\ 
    \noalign{\hrule height 1.5pt}
    \end{tabular}
    }
    \end{center}
    \vspace{-3mm}
    \caption{\textbf{Evaluations for rigid interactions} under HO3D~\cite{hampali2020honnotate} dataset.}
    \label{tab02_compare_rigid_Contact}
\end{table}


\begin{table}[t]
    \begin{center}
        \resizebox{1\linewidth}{!}{
    \begin{tabular}{c|cccc}
    \noalign{\hrule height 1.5pt}
    \rowcolor{white}Methods 
    & Initial State
    &~\cite{grady2021contactopt} &~\cite{tse2022s} &Ours
    \\  
    \midrule
    $\text{MPJPE}_{H}(\mathrm{mm})\downarrow$
    &18.54  &19.84 &19.76 &\textbf{18.97}    
        \\  
    $\text{MPVPE}_O(\mathrm{mm})\downarrow$ 
    &21.42   &21.45 &21.40 &\textbf{21.04}    
        \\  
    Max Pene.$(\mathrm{mm})\downarrow$ 
    &19.46    &13.19  &12.78 &\textbf{10.42}    
         \\  
    Inter.$(\mathrm{cm}^3)\downarrow$
    &14.25   &8.74   &9.02 &\textbf{8.56}     
        \\ 
    \noalign{\hrule height 1.5pt}
    \end{tabular}
    }
    \end{center}
    \vspace{-3mm}
    \caption{\textbf{Evaluations for nonrigid interactions} under HMDO~\cite{xie2023hmdo} dataset.}
    \label{tab03_compare_deform_Contact}
    \vspace{-4mm}
\end{table}




\subsection{Comparisons}

\begin{figure*}[!t]
    \centering
    \includegraphics[width=\linewidth]{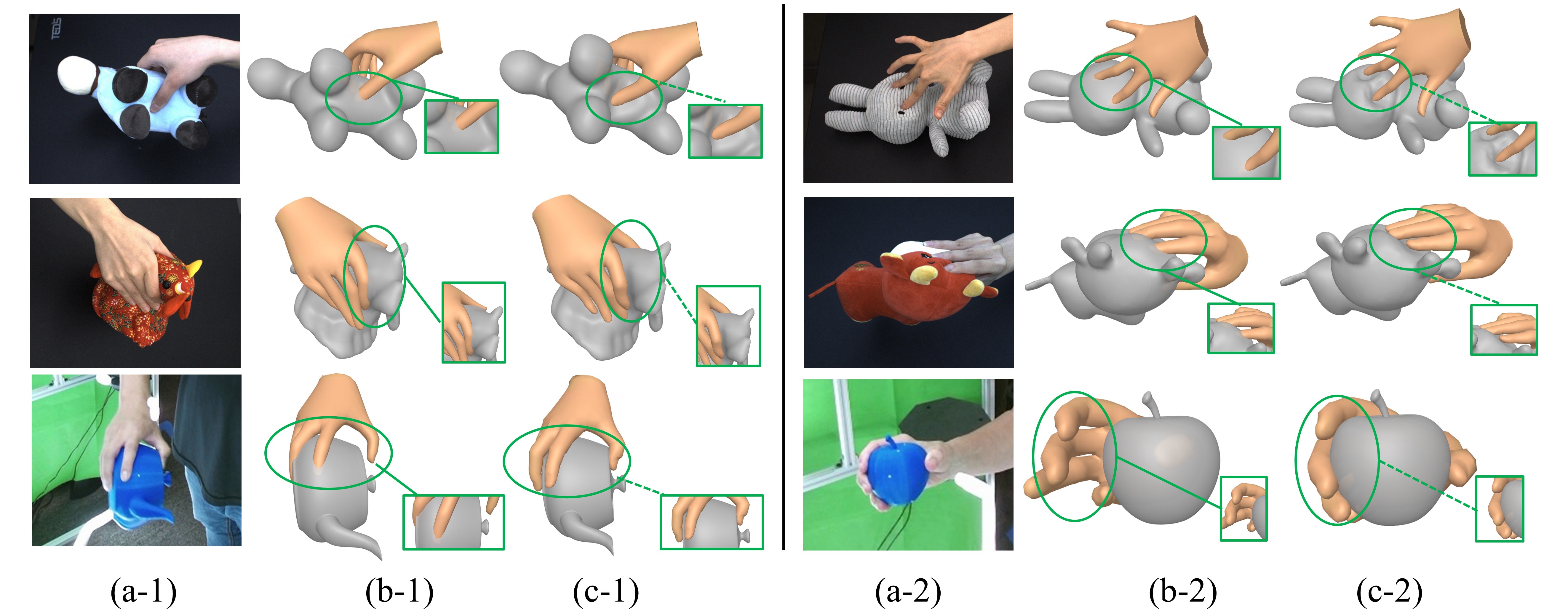}
    \caption{\textbf{More qualitative results.} (a) RGB images. (b) Initial hand-object surfaces. (c) Ours. High-quality reconstruction results certify the effectiveness of our framework.}
    \label{fig05_ours_rec}
\end{figure*}

\noindent\textbf{Monocular hand-object reconstruction. }In the task of reconstructing the hand-object from the monocular image, our method is compared with the hand-object reconstruction network from Hasson~\etal~\cite{hasson2020leveraging}. The quantitative results on HO3D~\cite{hampali2020honnotate} and FPHB~\cite{garcia2018first} datasets are shown in~\tablemk\ref{tab01_compareRecon}. Our method achieves better performance in hand-object interaction datasets. This demonstrates that our method can achieve explicit contact patterns inference and effective hand-object contact optimization, which can help us reconstruct hand-object interaction with higher quality. We show our qualitative results in~\figmk\ref{fig04_mono_rec}. Our methods can achieve more plausible reconstructions with fewer penetrations than~\cite{hasson2020leveraging}. More qualitative results of our method are shown in~\figmk\ref{fig05_ours_rec}.

\noindent\textbf{Hand-object contact estimation. }We take the result from our pose estimation network as the initial state and compare our \netname with ContactOpt~\cite{grady2021contactopt} and $\mathrm{S}^2$Contact~\cite{tse2022s}. 
We retrain DeepContact network in ContactOpt~\cite{grady2021contactopt} and GCN-Contact network in $\mathrm{S}^2$Contact~\cite{tse2022s} on the HMDO~\cite{xie2023hmdo} dataset.
As shown in~\figmk\ref{fig02_teaser} and~\figmk\ref{fig07_contact_driven}, we show the qualitative results compared with~\cite{grady2021contactopt, tse2022s} under HMDO~\cite{xie2023hmdo}, ContactPose~\cite{brahmbhatt2020contactpose} and HO3D~\cite{hampali2020honnotate} datasets. 
We evaluate the contact patterns optimization results of 3D rigid hand-object interaction between~\cite{grady2021contactopt,tse2022s} and ours in~\tablemk\ref{tab02_compare_rigid_Contact}. Our method achieved better performance than other methods. The contact optimization results of nonrigid hand-object interaction are shown in~\tablemk\ref{tab03_compare_deform_Contact}, and our method achieves higher quality grasping of hand and object. These demonstrate that our method can achieve better contact patterns optimization in both rigid and nonrigid interactions compared to other methods. Since \netname can estimate the deformed degree of the contact areas and the deformed transformations of the object, it allows our method to suitable for both contact optimization with nonrigid and rigid objects.
From~\tablemk\ref{tab02_compare_rigid_Contact} and ~\tablemk\ref{tab03_compare_deform_Contact}, it can be seen that although our method and~\cite{grady2021contactopt, tse2022s} did not improve hand pose estimation, they significantly reduced the intersection volume and max penetration, and improved the hand-object contact quality. This may be due to the optimization affected the hand region that did not interact with the object, as shown in the row 2 of~\figmk\ref{fig07_contact_driven}. 

\subsection{Ablation Study}
\label{Ablation Study}

\noindent\textbf{Baseline. } We take our hand-object pose estimation network as our baseline. Since monocular estimation is ill-posed, there may be mutual penetration or no contact between the reconstructed hand and the object. In addition, the contact areas of the object may be deformed due to its non-rigidity and the force exerted by the subject. Since HMDO~\cite{xie2023hmdo} is the dataset that records the 3D nonrigid interactions, most of our ablation experiments are based on this dataset. Where the results of completely using our entire pipeline are in the last row of~\tablemk\ref{tab04_ablation}.

\begin{table}[!t]
    \begin{center}
        \resizebox{1\linewidth}{!}{
    \begin{tabular}{c|cc|cc}
    \noalign{\hrule height 1.5pt}
    Datasets 
    &\multicolumn{2}{c|}{FPHB~\cite{garcia2018first}} 
    &\multicolumn{2}{c}{HO3D~\cite{hampali2020honnotate}} 
    \\
    \midrule
    \rowcolor{white}Methods 
    &~\cite{hasson2020leveraging}  & Ours  
    &~\cite{hasson2020leveraging}  & Ours  
    \\
    \midrule
    $\text{MPJPE}_{H}(\mathrm{mm})\downarrow$
        &18.23  &\textbf{17.86} 
        &14.74   &\textbf{14.78} 
       
        \\
    $\text{MPVPE}_O(\mathrm{mm})\downarrow$  
       &21.45   &\textbf{21.22} 
        &19.42  &\textbf{19.27} 
      
        \\
    Max Pene.$(\mathrm{mm})\downarrow$  
       &18.64  &\textbf{13.35} 
        &11.43  &\textbf{9.24} 
     
        \\
    Inter.$(\mathrm{cm}^3)\downarrow$
        &13.57  &\textbf{8.28} 
        &10.26  &\textbf{7.39} 
      
        \\
    \noalign{\hrule height 1.5pt}
    \end{tabular}
    }
    \end{center}
    \vspace{-2mm}
    \caption{\textbf{Comparisons for monocular reconstruction} under FPHB~\cite{garcia2018first} and HO3D~\cite{hampali2020honnotate} datasets.}
    \label{tab01_compareRecon}
    \vspace{-4mm}
\end{table}

\begin{table}
    \begin{center}
        \resizebox{1\linewidth}{!}{
          \begin{tabular}{c| cccc}
                \noalign{\hrule height 1.5pt}
                Method   & $\text{MPJPE}_{H}(\mathrm{mm})\downarrow$ &  $\text{MPVPE}_O(\mathrm{mm})\downarrow$  &Max Pene.$(\mathrm{mm})\downarrow$  & Inter.$(\mathrm{cm}^3)\downarrow$ \\
                \midrule
                Baseline  & 18.54 & 21.42 &19.46 &14.25 \\
                \midrule
               
                w/ RUP-32  & 19.12 & 21.16 &11.04 &8.82 \\
                w/ RUP-128 & 18.94 & 21.08 &10.25 &8.64 \\
              
                \midrule
                 w/o Con-Att& 19.73 & 21.59 &11.62 &9.27 \\
                 w/o Img-Pat& 19.96 & 21.65 &12.57 &9.19 \\
                \midrule
                 w/ Point-Tran& 19.04 & 21.09 &10.75 &8.49 \\
                 w/ grid-8 & 19.25 & 21.24 &11.16 &8.74\\
                \midrule
                Ours &  \textbf{18.97} & \textbf{21.04} & \textbf{10.42} & \textbf{8.56}\\
                \noalign{\hrule height 1.5pt}
          \end{tabular}
        }
    \end{center}
    \caption{\textbf{Ablation study of our method.} 
    Our \netname, surface unwrapping and point sampling are evaluated.}
    \label{tab04_ablation}
\end{table}

\noindent\textbf{Surface unwrapping. }We explore the effects of the size of hand-object RUPs. As shown in row 3 to row 4 of~\tablemk\ref{tab04_ablation}, we compared the impact of different sizes of RUP on hand-object interaction reconstruction. Considering both efficiency and reconstruction quality, it is appropriate to set the size of RUPs to $64\times 64$. RUP of $n \times n$ size is denoted as ``RUP-n'' in~\tablemk\ref{tab04_ablation}.


\noindent\textbf{Contact estimation. } The impact of contact attention modules is ablated as shown in row 5 of~\tablemk\ref{tab04_ablation}. We replace contact attention modules with an MLP architecture, resulting in a decrease in the quality of hand object reconstruction. The main reason may be that the contact attention modules can better explore the visual and hand-object spatial correlation, which can help better to predict the mutual contact areas and the object deformation. We ablate the effect of introducing image patches on the results. As shown in row 6 of~\tablemk\ref{tab04_ablation}, visual cues can improve the quality of hand-object reconstruction. Because the image patches contain contact and deformation information, which can guide our RUFormer to better predict contact areas and deformed degrees on region-aligned features. We denote the contact attention modules as ``Con-Att'' and image patches as ``Img-Pat'' in~\tablemk\ref{tab04_ablation}.

\noindent\textbf{Deformation estimation. } As shown in row 7 of~\tablemk\ref{tab04_ablation}, we ablate the deformation estimation block. We replace our deformation estimation block with Point-Transformer~\cite{zhao2021point}. Compared with~\cite{zhao2021point}, our deformation estimation block can give consideration to both efficiency and accuracy. We do not need to constantly query and build the neighborhood. Our deformation estimation block can benefit from the ordered sampling points and achieve effective aggregation of deformation features. In addition, we ablate the grid size for ordered point sampling from object-RUPs and object-f-RUPs, as shown in row 8 of~\tablemk\ref{tab04_ablation}. Since the deformed transformations aggregated from fewer grid-wise sampling features can not well represent the object surface deformation and more points calculations are expensive, setting grid size to $4 \times 4$ is more appropriate. We denote Point-Transformer as ``Point-Tran'' and $n \times n$ grid size as ``grid-$n$'' in~\tablemk\ref{tab04_ablation}.


\section{Conclusion} 
This paper proposes a learning-based framework to estimate the contact patterns between hand and nonrigid objects from monocular images.  A hand-object interaction representation is proposed to record the hand-object surfaces into multiple fine-grained 2D regional unwrapping profiles. Based on this representation, the roughly estimated hand-object surfaces are first unwrapped into 2D regional profiles, then a Vision Transformer is tamed to predict contact areas and deformed transformations within/across regions according to region-aligned features. Finally, hand-object surfaces are refined based on contact areas and deformed transformations.


\noindent\textbf{Limitations and Future Work.} Due to the influence of 2D hand joints on image patch cropping, our method relies on reliable 2D pose estimation. Our method can be extended to RGBD input and multi-view RGB input. By introducing depth and multi-view information, we can improve the quality of contact patterns and hand-object reconstruction.



{\small
\bibliographystyle{ieee_fullname}
\bibliography{egbib}
}

\end{document}